\documentclass{article}
\usepackage{ijcai23}

\usepackage{times}
\usepackage{soul}
\usepackage{url}
\usepackage[hidelinks]{hyperref}
\usepackage[utf8]{inputenc}
\usepackage[small]{caption}
\usepackage{graphicx}
\usepackage{amsmath}
\usepackage{amsthm}
\usepackage{booktabs}
\usepackage{algorithm}
\usepackage{algorithmic}
\usepackage[switch]{lineno}

\usepackage{todonotes}
\usepackage{booktabs}
\usepackage{nameref}
\usepackage{url}

\linenumbers

\title{Beyond Controls: Towards analyzing Communication choices in Mixed-Initiative Co-Creativity systems}
\author{Anonymized Authors}
\begin{document}

\maketitle

\maketitle

\begin{abstract}
    \begin{quote}

Generative Artificial Intelligence systems 
have been developed for image, code, story, and game generation with the goal of facilitating human creativity.
Recent work on neural generative systems has emphasized one particular means of interacting with AI systems: the user provides a specification, usually in the form of prompts, and the AI system generates the content.
However, there are other configurations of human and AI coordination, such as co-creativity (CC) in which both human and AI systems can contribute to content creation, and mixed-initiative (MI) in which both human and AI systems can initiate content changes.
In this paper, 
we define a hypothetical human-AI configuration {\em design space} consisting of different means for humans and AI systems to communicate creative intent to each other. 
We conduct a human participant study with 185 participants to understand how users want to interact with differently configured MI-CC systems. 
We find out that MI-CC systems with more extensive coverage of the design space are rated higher or on par on a variety of creative and goal-completion metrics, demonstrating that wider coverage of the design space can improve user experience and achievement when using the system;
Preference varies greatly between expertise groups, suggesting the development of adaptive, personalized MI-CC systems;
Participants identified new design space dimensions including scrutability---the ability to poke and prod at models---and explainability.

    \end{quote}
\end{abstract}

\section{Introduction}
The wider availability of generative AI systems in domains ranging from text~\cite{brown_language_2020}, image~\cite{ramesh_hierarchical_2022}, program code~\cite{chen_evaluating_2021}, 
to game stages and concepts~\cite{khalifa_pcgrl_2020},
is making the development of
creative content more accessible to people with more diverse backgrounds and skills.
Recent work on neural generative systems has emphasized one particular means of interacting with AI systems: the user provides a specification (e.g., prompt, previous text context, structured data, or one work of art to be restylized into another), and the AI system generates the content.
However,
the means of initialization, along with the majority of interactions between user and system,
are {\em not} human-centered.
In particular, they impose a specific paradigm of input on the human designer that best suits the underlying algorithms and models instead of the needs of the human designer.

Other configurations of human designer and AI creative system are possible that promise to reduce cognitive load, frustration, system abandonment~\cite{sweller_cognitive_2011}, and make these systems more casual and enjoyable \cite{compton2015casual}.
{\em Mixed initiative} (MI) systems are those in which both human and AI systems can initiate content changes.

{\em Co-Creative} (CC) systems are those in which both human and AI systems can contribute to content creation
Mixed initiative co-creative (MI-CC) systems have been demonstrated in game design~\cite{liapis_can_2016}, coding ({\em Github CoPilot}), drawing~\cite{davis_drawing_2015} and
storytelling~\cite{alvarez_story_2022}
Although the building of systems that make use of MI-CC traits may help us better understand how  users think and collaborate with creative AI systems, 
our understanding of the human factors that underly successful MI-CC systems remain relatively under-studied compared to the development of new
MI-CC systems.

In this paper,
we build on the dimensions of MI-CC systems identified by Lin et al.~\shortcite{lin_creative_2022}: Human vs. Agent-initiated, Elaboration vs. Reflection, Global vs. Local.
This framework defines a hypothetical design space for MI-CC systems where each value of each dimension can be instantiated as a specific way for a user to communicate creative intent with an AI system (and vice versa).
In the domain of story generation,
we conduct an exploratory human-participant study with 7 unique MI-CC systems as probes, each representing a plausible subset of the design space. \footnote{Code for the systems used in the study is available at \url{https://github.com/eilab-gt/beyond-prompts-experiment}}
We measure the perceived support the tool affords via the Creative Support Index (CSI)~\cite{cherry_quantifying_2014}.
Our study indicates that more extensive coverage of the design space can improve user experience and perceived creative achievement.
We also observe that preference for different types of communication with (and from) the system varies with expertise, suggesting the potential for adaptive, personalized MI-CC systems. 
Finally, our human-participant study uncovers a 4th dimension: explanation.

\begin{figure}[t]
\centering
\includegraphics[width=0.6\columnwidth]{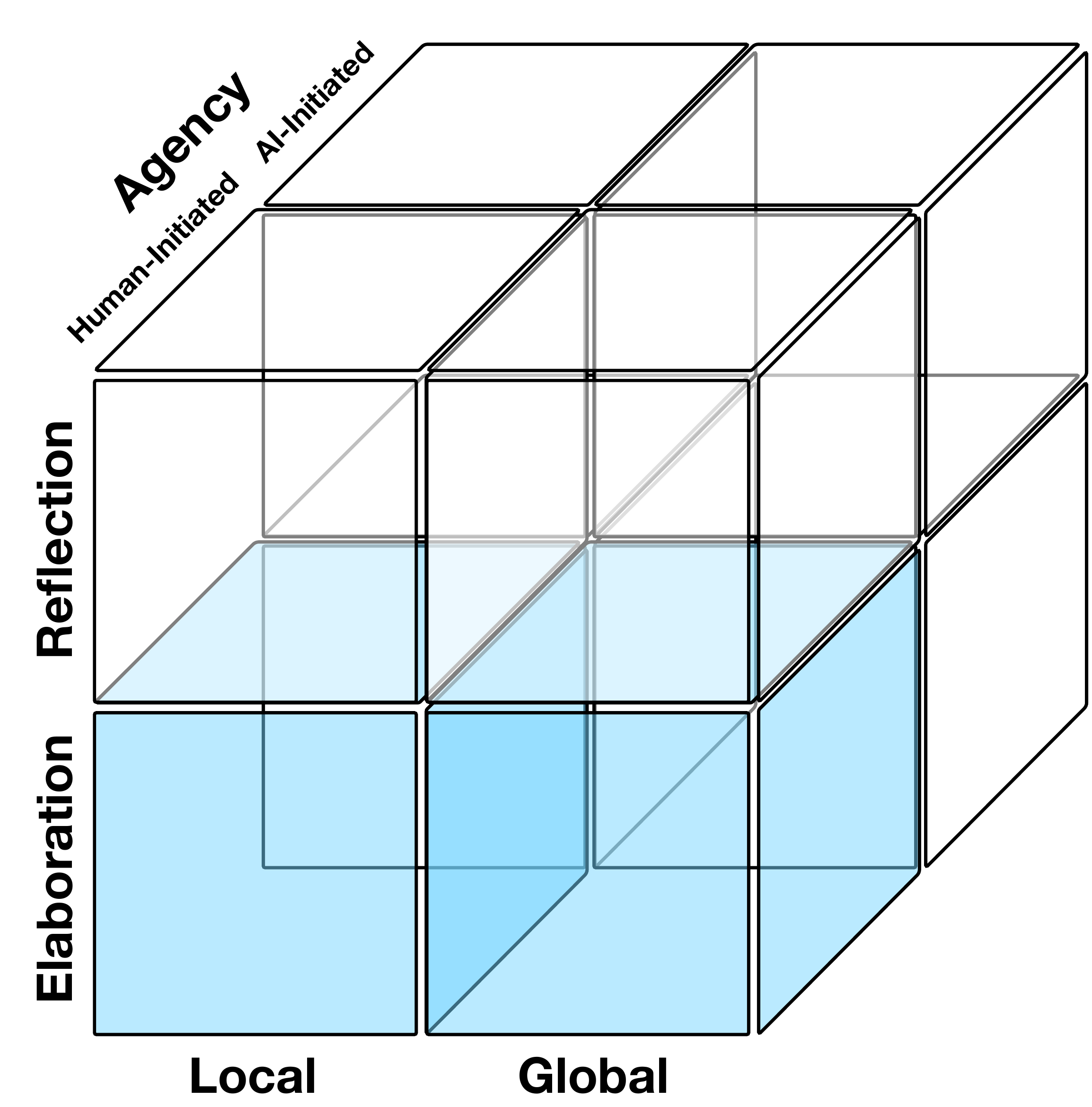}
\caption{The design space of user-AI communications in mixed-initiative co-creative systems considered in this work, consisting of three dimensions. 
The blue cubes have been explored in prior studies~\cite{lin_creative_2022}.
}
\label{fig:dimensions}
\end{figure}

\begin{figure}[t]
\centering
\includegraphics[width=1\columnwidth]{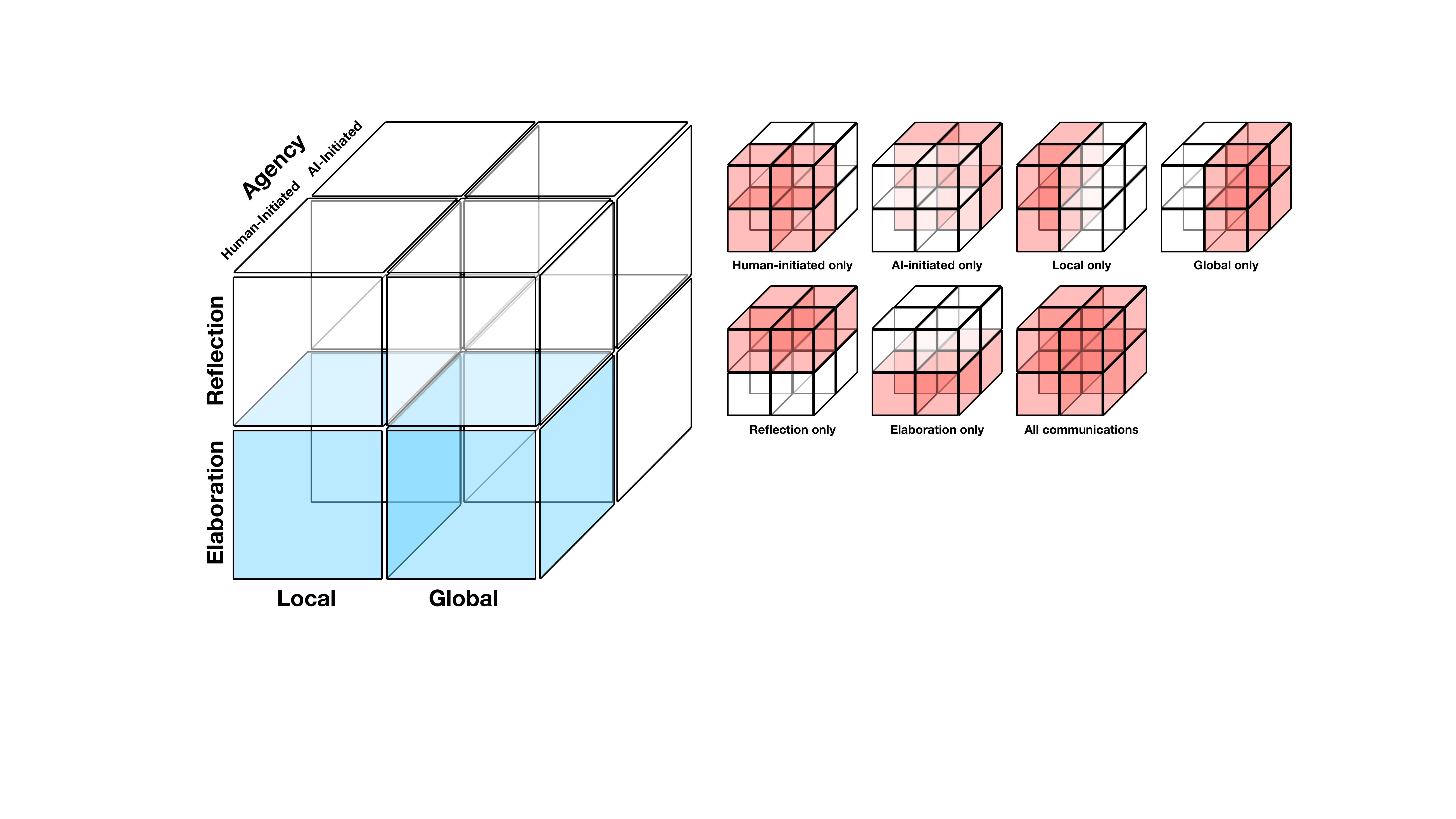}
\caption{The full system and six ablations---each removing one dimension from the design space---used as conditions in this work.
}
\label{fig:ablations}
\end{figure}

\section{Background and Related Work}
\label{subsection:related-micc}

A {\em mixed-initiative system} is one where ``a human initiative and a computational initiative" cooperate towards a shared goal~\cite{novick_what_1997}.
 In this work, we focus on storytelling in a game design setting.
However, as Lai, Leymarie, and Latham~\shortcite{lai_mixed-initiative_2022} point out, it can be easily applied to other creative domains.
Like that work we 
exclude so-called ``fire-and-forget'' systems~\cite{ramamurthy_is_2022} from consideration and focus on systems that allow iterative improvement on the creative artifact.

A {\em co-creative} agent~\cite{rezwana_cofi_2021,rezwana_designing_2022,guzdial_interaction_2019,grabe_towards_2022,kreminski_reflective_2021} is one that AI ``possesses the ability to alter the creative work equal to a human counterpart".
Note that when it comes to capability, ``ability" does not imply \textit{human-parity}.
Furthermore, as the human user and the AI system may tackle different parts of the creative work, it also does not entail equal \textit{responsibility};
Note that a mixed-initiative system does not necessarily need to be co-creative as the final product of the process does not have to be a creative artifact \cite{sekulic_evaluating_2022}.

To study the information that flows between the user and the MI-CC agent, 
we depart from the definition of Communications from \cite{lin_creative_2022},
which is itself based on works in categorizing or differentiating between different types of interactions between parties in an MI-CC system.
The framework by Rezwana and Maher~\shortcite{rezwana_cofi_2021,rezwana_designing_2022} that models interactions in co-creative systems that ``focus on flow of information between collaborators'', inspiring the definition of Communications.
Guzdial and Riedl~\shortcite{guzdial_interaction_2019} point out that human designers and AI ``can initiate the same action sets to modify the creative work, albeit with different executions''.
Kreminski~\shortcite{kreminski_reflective_2021} presents a survey on communications ``where the agent thinks about what happened in the process and takes actions based on it''.

Recent works on co-creative systems with a large array of interactive options include \textit{CoAuthor} \cite{lee_coauthor_2022}, wherein a text continuation setting fine-grained keyboard-based actions are recorded as a dataset, and 
\textit{CoPoet} \cite{chakrabarty_help_2022}, a poetry system wherein options are implemented as prompts constructed to represent requests with different configurations and ranges of application.
These works have shown the potential of co-creativity systems with a wide range of capabilities, further encouraging us on a comparative study over the design space of these capabilities, which is the focus of this work.

\section{Design Space for MI-CC Systems}
\label{subsection:communication}
\label{subsection:dimensions}

MI-CC systems can come in a variety of sizes and shapes.
Lin et al.~\shortcite{lin_creative_2022} presented a framework to help categorize them in terms of how the user and system communicate with each other and how information flows chronologically from one to the other.
The framework is domain agnostic, though demonstrated through a text generation example.
The framework contains three continuous, non-exhaustive dimensions to classify communications to and from user and system:
\begin{itemize}
    \item \textbf{Human-initiated vs. Agent-initiated}, which considers which of the two parties is initiating communication.
    \item \textbf{Elaboration vs. Reflection}, which deals with whether the communication relates to previously generated contents (reflection) or  newly planned actions (elaboration).
    \item \textbf{Global vs. Local}, which is based on the scope of the creative work of the communication.
\end{itemize}

We use the framework to guide the construction of variations of a MI-CC storytelling system.
By tying an axis $A$ to each dimension, the Cartesian product of these axes form a \textit{design space}. 
In the design space there are $2^3=8$ different ways that user and AI system can communicate.
For example, one means of communication might be human-initiated, involve elaboration of the creative artifact, and focus on local information.

The framework---and our work accordingly---considers {\em social} communication out of scope. 
Social communication is that between two creators that does not involve the sharing of information about the creative artifact or the alteration of it. 
Social communication, however, is not entirely unproductive as it may improve the relationship between human creator and AI agent, allow the user to better understand the AI system, and improve trust and rapport~\cite{margarido_lets_2022}.
The framework is also agnostic about the modality (text, visual, audio, etc.) of the communication, which Margarido et al.~\shortcite{margarido_lets_2022} also argues may be considered an additional dimension.

The framework and design space does not tell one {\em how} user-AI communication should occur beyond the broad properties provided such as who initiates, the extent of the information, and whether the information is about newly planned actions or previously made actions.
To that end, each type of communication can be implemented in many ways within a given MI-CC system. 
In the next section, we provide our specific instantiations of each of the eight types of communication.

Although conditional generation systems---GPT~\cite{brown_language_2020} for text, but also existing for other modalities (cf., \cite{ramesh_hierarchical_2022,khalifa_pcgrl_2020})---condition their generation on input such as a prompt from the user,
the choice of information exchanged among candidates is under-explored beyond so-called ``fire-and-forget'' systems which are trained on text corpus and/or task-based human feedback~\cite{ramamurthy_is_2022}.
Fire-and-forget systems, which can also be thought of as {\em assistants}, are a special case of the design space: human-initiated, elaborative, and local. 
Studying MI-CC systems with Communications from different points in the design space will help researchers to better understand how an AI system will collaborate with a human designer and facilitate generation systems to better align with tasks unique to MI-CC systems.

\section{Experimental System Overview}
\label{section:exp-sys}

To conduct an exploratory study of how the availability of different means of communication affect the actual and perceived creative experience, we used the {\em Creative Wand} framework~\cite{lin_creative_2022}, 
which is designed to facilitate experimentation with MI-CC systems.
The Creative Wand framework is a highly configurable MI-CC system made up of four abstracted components:
\begin{itemize}
    \item {\em Creative Context}: an interface between generative  algorithms related to the specific domain.
    \item {\em Experience Manager}: responsible for maintaining the state of the system.
    \item {\em Communications}: a set of modules that instantiate different means of communicating creative intent (see Section \nameref{subsection:comm_instantiation}.
    \item {\em Frontend}: defines how information is presented to, and received from, the user.
\end{itemize}

\subsection{Creative Context: Storytelling Domain and AI Algorithms}

Similar to Lin et al.~\shortcite{lin_creative_2022}, we
also consider textual story creation as the domain, situated as a key task in game development. 
In the story creation domain, the user attempts to create a plot with 10 lines.
Since it is a plot, the lines express the general activities of characters. 
See Figure~\ref{fig:screenshot} for an example.
Since story creation can be open-ended, we needed a way to constrain the activity in order to assess user performance.
To that end, we artificially provide the user with a goal.
We gave the same goal that \cite{lin_creative_2022} used, which is to create a story that starts with ``Business" and ends with ``Sports" while mentioning ``soccer".

In a MI-CC story creation task, the system must be capable of receiving communication from the human designer about creative intent at various levels (global, local), but also providing critical reflection on the story content.
As there is no one AI story generation system capable of doing everything we need for all aspects of the design space,
We deployed two existing AI systems:
{\em Plug and Blend}~\cite{lin_plug-and-blend_2021} updated to use the larger GPT-J~\cite{wang_gpt-j-6b_2021}  language model instead of GPT-2~\cite{radford_language_2019};
and
{\em CARP}~\cite{matiana_cut_2021}.

\subsubsection{Plug and Blend}
This system uses two models to generate text that adheres to topic controls.
The first model is a standard, unaltered large language model.
In the case of this paper, we use the GPT-J pre-trained large language model, which accepts a prompt or context text that begins the story.
The second model learns a set of weights that can be applied to the output logits of the language model output in order to bias the generation toward a particular topic.
A set of topics and the sentence spans they should be applied to is provided as a second type of input to Plug and Blend,
referred to as a {\em sketch};
They are translated to individual control strength that amplifies the weight applied from the second model, and further guides the generation of paragraphs.
We modified the pipeline so that ``Story for kids: Once upon a time,'', concatenated with at most two previous lines, as the prompt, along with the topic control, is used to generate each line of the story in our system.

\subsubsection{CARP}
We used the CARP model \cite{matiana_cut_2021}, a language model that is trained on contrastive objectives to learn a cosine similarity score between a sentence and a short critique, such as ``This character is confusing''.
CARP cannot generate narratives, but can score individual lines in a narrative according to a given criterion.
It is the basis for our communication modules involving reflection.
CARP produces values between $[0.15,0.4]$, which we rescale to $[0,1]$.

\subsection{Communications}
\label{subsection:comm_instantiation}

We designed 11 modules for communication to cover the entirety of the design space, as well as some additional communication modules that emulate basic functionality that many expect in creation tools.
We give the 11 communication modules below, indicating where on each of the three axes it falls.
We organize the list around the axis of elaboration vs reflection because elaboration and reflection are tied to the two AI algorithms.

\subsubsection{Elaboration Communication Modules}

Elaboration communications are related to generation of new contents, and use the Plug and Blend AI algorithm.
To batch user input for a better user experience, we do not immediately start the regeneration process until the user requests the story to be rewritten, which is one of the three miscellaneous communications.

\begin{itemize}

    \item 
    {\bf Write a sentence} 
    \scalebox{0.75}{$\left[\substack{Elab.\\Human\\Local}\right]$}
    The user provides a specific sentence to be inserted at a particular line index.
    If there is already text in that line, it gets replaced. 

    \textit{Example:} The user replaces the first line with "Hello!".
    
    \item 
    {\bf Apply a topic}  
    \scalebox{0.75}{$\left[\substack{Elab.\\Human\\Global}\right]$}
    The user provides a topic code along with a starting line index and an ending index.
    We provide four pre-defined options---``business'', ``sports'', ``science``, ``world``.
    The user can type in their own free text code as well.
    The Plug and Blend sketch data structure is updated though generation does not happen until the user requests re-generation.
    
    \textit{Example:} The user applies ``Business'' to the first five lines of the story.

    \item {\bf Get a sentence suggestion} 
    \scalebox{0.75}{$\left[\substack{Elab.\\Agent\\Local}\right]$}
    The Plug and Blend AI algorithm generates a new candidate sentence based on a random existing line and a topic chosen between ``Business'' and ``Sports'', focusing in on the two goals participants are asked to meet.
    The user can then choose to accept or reject the suggestion.

    \textit{Example:} The agent provided ``Football is interesting...'' as a suggestion to line 3 of the story.

    \item 
    {\bf Get a topic suggestion} 
    \scalebox{0.75}{$\left[\substack{Elab.\\Agent\\Global}\right]$}
    The user is provided a topic suggestion between ``Business'' and ``Sports'' chosen randomly.
    If the user decides to accept the suggestion, they continue the process as in applying a topic control.
    
    \textit{Example:} The agent provides ``Sports'' as the suggested topic.
    The user accepted the topic and decided to apply it to the last 5 lines of the story.

\end{itemize}

\subsubsection{Reflection Communcation Modules}

This family of communications uses the CARP model, which pairs a critique and each sentence with a score signifying how related they are.
For each communication below, this information is used differently, but all towards providing the user with insights or opportunities to think about whether the story so far needs further modification.
Being provided with a critique, the system highlights lines of the story with increasingly brighter color relative to the rescaled score.
\begin{itemize}

    \item 
    {\bf Off-topic checker} 
    \scalebox{0.75}{$\left[\substack{Reflect\\Human\\Local}\right]$}
    The user picks a sentence in the story and gives a topic, and the agent tells the user whether it's related to that topic.
    We used ``This part of the story should be related to $\langle input\rangle$'' as the critique for the CARP model.

    \textit{Example:} The user selects line 3 which says ``Football is interesting...'' and asked whether it is related to ``Science''.
    
    \item 
    {\bf Reflect together}  
    \scalebox{0.75}{$\left[\substack{Reflect\\Human\\Global}\right]$}
    The user gives a critique, and the agent highlights sentences based on the score given from the CARP model.

    \textit{Example:} The user provides ``It should be raining'', and the agent highlights a line in the story that says ``It's a sunny day'' (as well as any other lines based on how much they fail to match the critique). 
    
    \item 
    {\bf Get a local story quality tip}
    \scalebox{0.75}{$\left[\substack{Reflect\\Agent\\Local}\right]$}
    The agent picks a tip from a list of pre-determined critique prompts and highlights sentences based on the scores based on how much the match the critique according to CARP.

    \textit{Example:} The agent picks the ``The story should be fun'' pre-defined critique and highlights line 7 of the story, showing that the agent thinks that line is fun.
    
    \item 
    {\bf Get a high-level story quality tip} 
    \scalebox{0.75}{$\left[\substack{Reflect\\Agent\\Global}\right]$}
    The agent picks a tip from the same pre-defined set of critique prompts and tells the user whether the story as a whole is related to it.

    \textit{Example:} The agent picks ``Whether the story is about Sports'' from its set of critiques and tells the user ``yes'', confirming the user that the story is about ``sports''.

\end{itemize}

\subsubsection{Miscellaneous Communication Modules}

We also included three additional communication modules to fill in functionality of the system.

\begin{itemize}
    \item 
    {\bf Write the whole story} 
    When selected the system will re-generate all lines in the story, given a context prompt and the Plug and Blend sketch.
    This communication can be used multiple times in succession for alternatives.

    \item 
    {\bf Rewrite from a specific line} 
    Instead of starting fresh, the system only generates lines for the story after a specified point, leaving previous lines intact.
    This can only be used if a story already exists.

    \item 
    {\bf Undo} The system reverts the last operation.

\end{itemize}

\subsection{Experience Manager and Frontend}

\begin{figure*}
    \centering
    \fbox{\includegraphics[width=0.8\textwidth]{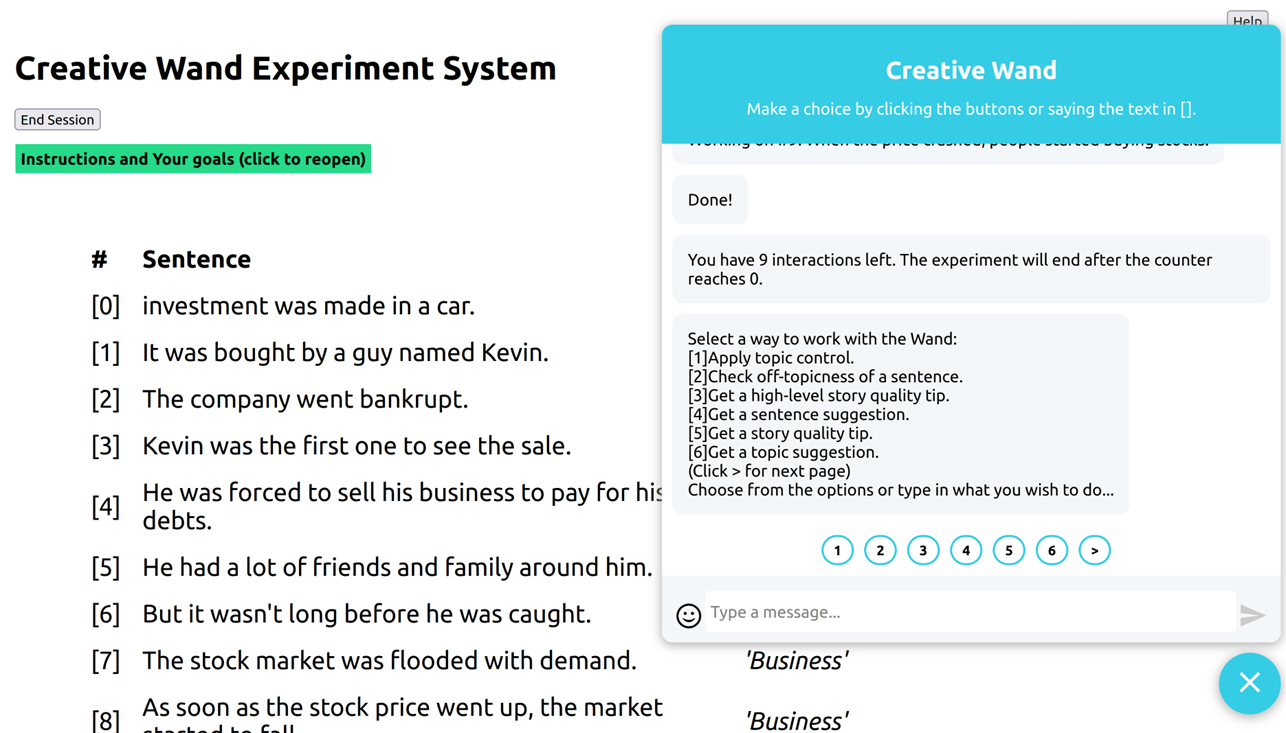}}
    \caption{Screenshot of our experiment system, along with instructions.}
    \label{fig:screenshot}
\end{figure*}

We implement a turn-based experience manager based on the sample Creative Wand implementation in Lin et al.~\shortcite{lin_creative_2022}.
For each turn, the manager provides available options for the user in a chatbot-like dialogue box, each of which maps to a communication module (for agent-initiated communications, entry point to allow the agent to take the initiative.) %
See Figure~\ref{fig:screenshot}, which shows a portion of the user interface.
We extend the Creative Wand framework to provide user experience by enhancements such as reverting back to previous states (i.e., undo).

\section{Study Methodology}
\label{section:survey-method}
To study the design space of communications in MI-CC systems, we developed seven versions of the story creation system described in the previous section.
One version had communication modules from every part of the design space.
The other six versions removed communications along a single dimension.
See Figure~\ref{fig:ablations}.

In this exploratory study we seek to determine 
how the presence or absence of different modes of human-AI communication affect perceptions of creative support.
We also seek to determine if individual variables such as creative background and familiarity with AI affects the above.

We recruited 185 participants on Prolific\footnote{\url{prolific.co}} who were automatically screened by the platform for adequate English proficiency.
Each experiment session lasts for approximately 30 minutes, and we paid the participants \$15 per hour.

Participants are first asked to complete a questionnaire about their creative background and familiarity with AI.
There were three multiple choice questions, as given below with their possible choices:
\begin{enumerate}[left=0pt, label=\Alph*.]
    \item Level of confidence with using a computer to author contents:
    \begin{enumerate}[left=0pt, label=A\arabic*.]
        \item I do not use computers to create things.
        \item I have used computers to create things, but for the past year, I have not done it once a week.
        \item I use computers to create things more than once a week, but I'm doing it not for the job (for example, for interest).
        \item I use computers to create things for my job.
    \end{enumerate}
    \item Level of confidence with using a computer to create games:
    \begin{enumerate}[left=0pt, label=B\arabic*.]
        \item I never used a computer to create anything related to games.
        \item I've done some work in the realm of games, but for the past year, I have not done it once a week.
        \item I create content for games out of interest, for more than once a week.
        \item I create content for games for my job.
    \end{enumerate}
    \item Familiarity with AI:
    \begin{enumerate}[left=0pt, label=C\arabic*.]
        \item All I know is no more than it being a buzz word.
        \item I have experience using something with "AI technologies" with it.
        \item I understand how recent AI technologies work.
    \end{enumerate}
\end{enumerate}

We then show instructions to familiarize them with how to use the experiment system.
This consists of annotated screenshots of the interface during different stages of the study, and a brief introduction to the workflow of co-creating a story.\footnote{Individual communications are not included in the tutorial, but further instructions are triggered the first time each communication is activated.}

Participants are then assigned to a random condition in which they will interact with two versions of the system: the ``full'' system (all communication modules), and one of the six ablations (communication modules removed in one dimension).
We counter-balanced the order we present the systems so that participants randomly start using either "full" or the ablation we assigned.

For both systems, participants were asked to create a story that was 10 lines long, starts with the topic of ``business'', ends with the topic of ``sports'', and mentions ``soccer'' at least once.
This is the same task and goal criteria as used by Lin et al.~\shortcite{lin_creative_2022}.
Participants are given 12 interactions with each system. An interaction is only complete once the participant provides all information for the system to execute the option and doesn't change or cancel the communication.%

\begin{table*}[!ht]
    \centering
    \scriptsize
    \begin{tabular}{p{0.15\linewidth}|c|cccccc}
         & \textbf{Overall} & \textbf{Agent-Init. Only} & \textbf{Human-Init. Only} & \textbf{Elaboration Only} & \textbf{Reflection Only} & \textbf{Global Only} & \textbf{Local Only} \\ \hline
        Num. valid responses & 185 & 31 & 32 & 30 & 32 & 27 & 33 \\ \hline
        Q1: Expressiveness & 62.2\%* & 74.2\%* & 46.9\% & 56.7\% & 78.1\%* & 63.0\% & 54.5\% \\ \hline
        Q2: Enjoyment & 60.5\%* & 74.2\%* & 43.8\% & 50.0\% & 81.2\%* & 59.3\% & 54.5\% \\ \hline
        Q3: Exploration & 62.7\%* & 71.0\%* & 46.9\% & 56.7\% & 71.9\%* & 70.4\%* & 60.6\% \\ \hline
        Q4: Immersion & 62.2\%* & 71.0\%* & 50.0\% & 60.0\% & 75.0\%* & 59.3\% & 57.6\% \\ \hline
        Q5: Collaboration & 59.5\%* & 71.0\%* & 40.6\% & 56.7\% & 81.2\%* & 59.3\% & 48.5\% \\ \hline
        Q6: Result worth effort & 60.5\%* & 64.5\%+ & 53.1\% & 60.0\% & 71.9\%* & 66.7\%+ & 48.5\% \\ \hline
        Q7: Better responses & 61.6\%* & 67.7\%* & 56.2\% & 63.3\% & 78.1\%* & 59.3\% & 45.5\% \\ \hline
    \end{tabular}
    \caption{Rate of participants that preferred the Full System over the ablations. * represents a significance level of $p<0.05$ on Full system preferred over the ablation; + for $p<0.1$. No ablation was preferred statistically significantly.}
    \label{table:ablation}
\end{table*}

Once they finished using both systems, as the exit survey, participants were asked to complete another questionnaire with seven questions about their satisfaction with the process and the generated story.
These questions are presented in random order for each participant.
The first six questions are adapted from the Creative Support Index (CSI)~\cite{cherry_quantifying_2014}, which is a validated measure of how well a tool supports human creativity.
We ask: between the ``full'' system and the ablation, ``Which system is more associated with each of the statements?'':
\begin{enumerate}[left=0pt, label=Q\arabic*.]
    \item \textbf{(Expressiveness)} This system made it easiest for me to express and share my goals, given to me in instructions, with the AI system.
    \item \textbf{(Enjoyment)} I enjoyed interacting with this system most.
    \item \textbf{(Exploration)} This system was most helpful for exploring different ideas and possibilities.
    \item \textbf{(Immersion)} This system made me feel the most absorbed in the task to the point that I forgot I was working with the system.
    \item \textbf{(Collaboration)} This system best allowed me to achieve the goal assigned to me.
    \item \textbf{(Results worth effort)} This system provides the overall best quality stories by the time I was done.
\end{enumerate}
Additionally, we also asked:
\begin{enumerate}[resume*, left=0pt, label=Q\arabic*.]
    \item Which system tends to get the best response for the same type of requests?
\end{enumerate}
We anticipate no preference between both systems on Q7, as the implementations of how the systems handle these requests (provided that an ablation system has that capacity) is unchanged.
As an attention mechanism, we also asked what the perceived similarity and differences between these two systems are before the participant finishes the study.

\section{Results and Discussions}

\paragraph{Participant Creative Background}
98\% of the participants reported that they at least used computers to create things (A2-A4), and 41\% say they do it as their job (A4).
Although we did not specifically recruit people with experience in designing game contents,  
49\% of the participants identify them as at least carrying out some work in the realm of games (B2-B4).
84\% reported that they have used something with AI technologies (C2-C3), and 26\% say they know how recent AI works (C3).
Table~\ref{table:q_vs_demo} (column 1-2) shows how many participants responded yes to each question.

\begin{table*}[!ht]
    \centering
    \footnotesize
    \begin{tabular}{|c|c||c|c|c|c|c|c|c||c|c|c|}
    \hline
        & \textbf{$n$} & \textbf{Q1} & \textbf{Q2} & \textbf{Q3} & \textbf{Q4} & \textbf{Q5} & \textbf{Q6} & \textbf{Q7} & \multicolumn{3}{|c|}{\bf Welch's $t$-test} \\ \hline\hline
        \multicolumn{9}{|c||}{\bf Level of confidence with using a computer to author contents (A1-A4)} & {\bf A2} & {\bf A3} & {\bf A4} \\ \hline
        \textbf{A1} & 3 & \multicolumn{10}{|c|}{\em Too few participants} \\ \hline
        \textbf{A2} & 44 & 61.4\% & 56.8\% & 54.5\% & 54.5\% & 61.4\% & 56.8\% & 56.8\%&N/A&*&\\ \hline
        \textbf{A3} & 62 & 66.1\% & 67.7\% & 72.6\% & 67.7\% & 66.1\% & 69.4\% & 71.0\% &*&N/A&*\\ \hline
        \textbf{A4} & 76 & 60.5\% & 57.9\% & 60.5\% & 63.2\% & 53.9\% & 56.6\% & 57.9\% & &*&N/A\\ \hline\hline
        \multicolumn{9}{|c||}{\bf Level of confidence with using a computer to create games (B1-B4)}  & {\bf B1} & {\bf B2} & {\bf B3}\\ \hline
        \textbf{B1} & 94 & 63.8\% & 63.8\% & 60.6\% & 66.0\% & 64.9\% & 67.0\% & 66.0\% &N/A&*&\\ \hline
        \textbf{B2} & 65 & 52.3\% & 50.8\% & 66.2\% & 52.3\% & 46.2\% & 49.2\% & 53.8\% &*&N/A&*\\ \hline
        \textbf{B3} & 21 & 81.0\% & 76.2\% & 61.9\% & 76.2\% & 71.4\% & 61.9\% & 61.9\% & &*&N/A \\ \hline
        \textbf{B4} & 5 & \multicolumn{10}{|c|}{\em Too few participants} \\ \hline\hline
        \multicolumn{9}{|c||}{\bf  Familiarity with AI (C1-C3)} & {\bf C1} & {\bf C2} & {\bf C3}\\ \hline
        \textbf{C1} & 29 & 48.3\% & 55.2\% & 48.3\% & 51.7\% & 48.3\% & 58.6\% & 58.6\% &N/A&*&+\\ \hline
        \textbf{C2} & 107 & 66.4\% & 64.5\% & 67.3\% & 67.3\% & 61.7\% & 58.9\% & 63.6\% &*&N/A&\\ \hline
        \textbf{C3} & 49 & 61.2\% & 55.1\% & 61.2\% & 57.1\% & 61.2\% & 65.3\% & 59.2\% &+& &N/A\\ \hline
    \end{tabular}
    \caption{Rate of preference on Full system, grouped by answers to demographics questions. Only data for groups with more than 20 participants are shown. * means different distribution with $p<0.01$, + for $p<0.1$. }
    \label{table:q_vs_demo}
\end{table*}

\paragraph{Perceptions of Creativity Support}
Table~\ref{table:ablation} shows the preference of users between the ``full'' system and the ablations
on the seven questions Q1--Q7.
Participants prefer the full system overall. 
When considering only the Agent-Initiated and Reflection ablations, the preference for the full version is also statistically significantly preferred on all questions.
That is: removing human-initiated communication or elaboration communication significantly degraded the creative experience in every measurement.
The ability to fully or partially generate the story was always an option.

The Global-only ablation, which removed communications involving local changes, was significantly less preferred than the full version only when considering the questions on exploration, and ``results worth it''. 
This suggests that global communications were not sufficient alone for exploring different ideas and participants felt less overall satisfaction with their story results when unable to make localized changes. 
Even though in many cases the full version was preferred over other ablations more than 60\% of the time, when participants are spread across conditions, there is a higher bar for statistical significance.

Participants were the most indifferent when comparing the full system to Local and Human-Initiated ablations, removing Global and Agent-initiated communications, respectively.
That is, removing these resulted in less reported degradation of the creative experience. 
In the Human-Initiated ablation, the AI is the most passive and never does anything until users provide enough information for them.
Most non-MI-CC systems operate this way and may be used at least weekly by 75\% of the participants (A3-A4).
Global communications are likely harder to use than local communications.
Participants were asked to learn a new creative support tool in a less-than-30-minutes experiment, with a sharp learning curve toward mechanisms that the participants are not familiar with in the first place.
These might have played a role.
Although a follow-up longitudinal study may help investigate this effect and provide a more accurate picture to study \textit{specific parts of the design space},
in our opinion, this also hints that not all dimensions in the design space of communications
are of equal value to users.

The full system was not significantly preferred over the Elaboration-only ablation, even though the full version was preferred 56\%--63\% of the time. 
This suggests that participants were more sensitive to the loss of reflective communications than the loss of agent-initiated or global communications. 
The role of reflective communication deserves further study; 
this study cannot determine the extent to which the specific use of CARP as the model for processing story critiques play in participant perception of reflective communications.

Surprisingly, participants prefer the ``full'' system on Q7---the system provides better responses---despite the fact that the AI systems were the same across all systems (when not removed due to the elaboration-only or reflection-only ablations). 
We hypothesize that because the "full" system can help the users achieve the goal better (61.6\% with statistical significance), it is likely that intermediate stories are also easier to work with; the communications ``give the better response'' because they have a better story to work on.

The overall trend is that \textbf{a wider coverage of the design space of user-AI communication types is appreciated}.
The study provides in-depth understanding of the relative significance of different types of user-AI communication.
The fact that our study cannot distinguish a preference for Local and Global communications suggests that they may both be important (except where noted otherwise above).
This is notable due to the absence of global communication modes in most ``fire and forget'' systems.

\paragraph{Individual Differences}
For each background experience question (A, B, C), we place participants in separate groups based on which multiple-choice option they selected.
This creates groups A1 through A4, B1 through B4, and C1 through C3.
Table \ref{table:q_vs_demo} shows the number of participants sorted into each group and how they responded to the creativity support questions.
We excluded groups with less than 20 participants from the analysis.

We conduct a Welch's $t$-tests between all groups and observe strong significant differences ($p<0.01$) between Groups A2 and A3, A3 and A4, B1 and B2, B2 and B3, and C1 and C2 with regard to how each group responded to the creativity support questions.
Except for C3 in which we observe a weaker difference ($p<0.1$) between C1.
Participants from each expertise group differ in their preference for the full system from at least one other group, suggesting that {\bf MI-CC tools should be customized to different types of users with different levels of creative expertise and AI familiarity}.

\begin{figure}[t]
    \centering
    \includegraphics[width=0.8\linewidth]{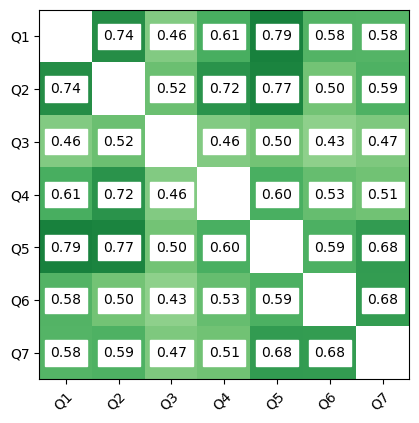}
    \caption{Correlation between questions in the survey.}
    \label{fig:correlation}
\end{figure}

\paragraph{Creativity Support Index Questions are Correlated}
We conduct a correlation analysis on the questions asked in the survey
(Figure~\ref{fig:correlation}).
We observe a medium (0.43) to strong (0.79) correlation between the six questions adapted from the Creativity Support Index~\cite{cherry_quantifying_2014}.
While we would expect all questions to correlate with each other because they are all, at some level, measuring different dimensions of creativity support, the data reflects participant response to the presence or absence of communication types. 
We see Q1 (expressiveness) and Q2 (enjoyment) and Q5 (collaboration) correlate at $>0.7$, suggesting the factors impact user perceptions of expressiveness are the same that impact perceptions of enjoyment.
Likewise we see Q2 (enjoyment), Q4 (immersion) correlate at $>0.7$, suggesting that the factors that impact perceptions of joy are the same that impact perceptions of immersion.
Due to the nature of the task, these factors are the presence or absence of certain communication modes, though we do not have fine-grained detail enough to identify which ones.
This further suggests that {\bf the communications that make the creative experience enjoyable are the same as those that make the experience seem immersive, expressive, and collaborative}.

Q7 (better responses) is strongly correlated with Q6 (results worth it).
Q7 was not derived from the Creativity Support Index, but this correlation provides further explanation for the observations about Q7 earlier that the perceived quality of AI responses would be correlated with perceptions of satisfaction with the creative outcome.

\subsection{Qualitative Findings \& Discussion}

We also analyzed open-ended justifications participants provided for their perceived levels of satisfaction with the systems using approaches inspired by thematic analysis~\cite{aronson1994pragmatic}. Taking an inductive approach, we started the process with an open-coding scheme and iteratively produced in-vivo codes (generating codes directly from the data). Next, we analyzed the data using axial codes, which involves finding relationships between the open codes and clustering them into different emergent themes. Through an iterative process performed until consensus was reached, we share the most salient themes below. 

Participants valued the \textbf{ability to exercise control} over the co-creative writing tool. Whichever tool was ``easier to control the topic'' (P09) was often favored. Customizability was a prized asset-- they felt that the customizability allowed ``the story to go together'' and be more coherent (P98). This notion of controllability was also associated to the tool’s ability to ``take topics into consideration'' (P64), which indicates how participants ascribed comprehension abilities in the tool based on their ability to control it. The sentiment is expressed succinctly by the following participant:
\begin{quote}
    Fish\footnote{Codename for Full system} 
    is superior to Rabbit\footnote{Codename for the Reflection-only ablation.} 
    in that it you can guide and interact with it and \textit{it listens to feedback} and doesn't just write what it wants. Fish allowed you more control in guiding the story on topics before starting so it was more accurate and also more customizable. Rabbit felt more random with less options and control, it started off topic and stayed off topic even when being prompted. Fish overall was a lot better than Rabbit. (P29, emphasis added)
\end{quote}

There was a \textbf{desire for scrutability}---to poke and prod---to \textbf{get a mechanistic or functional understanding} of the tools. This theme follows from the previous theme around the desire to control. Participants exhibited a desire to ``understand the mechanism of checking how one sentence is related to a particular topic'' (P07). The more control a tool allowed, the higher its perceived scrutability. Participants were trying to achieve a mechanistic understanding~\cite{lombrozo2019mechanistic}---how things worked—as well as functional understanding~\cite{lombrozo2014explanation} of the “why” behind the actions. A major part of this understanding was \textit{reciprocal} and \textit{mutual}; that is, participants felt that they could understand the tool if the tool could understand their instructions or input: 
\begin{quote}
    I had an easier time understanding the Fish system. And it appeared to understand the topics better based on my interaction. (P76)
\end{quote}

A core implication of both of these findings around control and scrutability suggest that adding explainability to these systems can enable argumentations, expose creative processes and augment the user’s mental model  \cite{llano_explainable_2022} and thereby foster better collaboration. Emerging work in Explainable AI (XAI) showcases that user backgrounds matter—that is, who opens the “black-box” matters when it comes to making sense of the AI’s output~\cite{ehsan2021explainable}. This entails that we need to customize the explainability according to the user’s background (which can include AI literacy, levels of experience, and familiarity). Moreover, we can also fine-tune explanations that target specific types of understanding such as mechanistic or functional. Each type of understanding is goal dependent; therefore, the explanations also have to be appropriately actionable~\cite{ehsan2021expanding}.

\section{Limitations}

While we aim at studying the design space of communications by picking up ones that best express their  neighbourhood, the three dimensions we borrowed from Lin et al.~\shortcite{lin_creative_2022} is not complete;
communications can also feature traits from both sides of an axis, such as ``adding details to an existing sentence" being both elaborative and reflective.
As we focus on \textbf{what} information is passed between the user and the AI agent, we controlled all the systems we used in the experiments to use the same User Interface and limited representation of text and highlight colors.
We invite colleagues alike to conduct similar experiments on other dimensions of Communications and representations, potentially in other modalities (image, speech, and more).

Arguably, agent-initiated communications still need human users' approval to initiate, as the particular implementation requires all communications to be triggered by the user selecting an option in the menu.
We made this decision to unify the representations of the Communications in the study.
Although we argue that the capability of the agent selecting which communication to trigger \textit{actively}, which is ultimately a decision-making problem over all communications, a topic that is beyond the scope of this work,
Since the goal of our work is to study MI-CC systems, we decided to pick a generative system that strikes a balance between availability and consistency with regard to the MI-CC experience we need for the study.

\section{Discussions}
Generative language models (LMs) rapidly advance - While this paper was being reviewed, ChatGPT, GPT-4\footnote{chat.openai.com} and a family of large-scale LMs that utilizes RLHF~\cite{ramamurthy_is_2022} demonstrated to the whole world end-to-end capabilities for collaborative authoring, where a dialogue agent can both generate contents based on an initial prompt, and amend what is just generated with follow-ups, all provided by human users in natural languages.
The usage of RLHF, where a reward model is trained to forecast human preference for the dialogues and then used to influence what LMs generate, is a crucial asset of these systems with regard to MI-CC. 
However, such systems are nominally mixed-initiative, as they by design is a question-answering and continuation assistant, by design only providing \textit{post-hoc} \footnote{\textit{Post-hoc} prompt-based explanations answer "Based on the decision I \textit{already} made, why?". Due to their token-based probabilistic continuation nature they are not designed to give \textit{ad-hoc} explanations of how they made decisions.} contents and explanations when user requests them;
They also have spaces for improvements as a co-creator, as RLHF-enhanced LMs rely solely on the context (LM) and ``mean" preferences of \textit{a sampling of general public} (RLHF), which is insufficient as we already demonstrated in this work that at least user-specific preferences and their prior experience also plays a role.

We believe, with these LMs and alike showing generation capability \textit{when the prompts are right}, the golden age of studying MI-CC systems has arrived: \textbf{Beyond prompts}, MI-CC systems that stands on these new frameworks have the potential to learn how to collaborate with specific users and truly co-create contents without the cognitive load of prompt engineering and procedures alike.
``Instead of a model teaching \textit{you} how to work with it, you should teach and collaborate with it." We leave this as future work.

\section{Conclusions}
We present a comparative study with 185 participants on MI-CC systems
that only differs in their inclusion or exclusion of particular modes of user-AI communication.
We find a trend that MI-CC systems with a wider coverage of user-AI communication types is appreciated,
and that preference also varies greatly between expertise groups, suggesting for the development of customized MI-CC systems for different types of users.
Participants also exhibited a desire for scrutability-- to poke and prod--to develop a mechanistic and functional understanding of the system where explanations can be useful.

Based on this evidence, we recommend that
designers of MI-CC systems should pay attention to the \textit{design space} of user-AI communications,
carefully study their audience, 
and plan for adaptation of their system towards individual users,
when sketching the interaction paradigm.
These insights can facilitate further MI-CC research, and, most importantly, encourage tailored collaborative experience for each designer (of diverse experience levels) to achieve their potential during co-creativity as well as the final output of the process.

\bibliographystyle{iccc}
\bibliography{references,iccc}

\end{document}